\documentclass[conference]{IEEEtran}
\usepackage{cite}
\usepackage{graphicx}
\usepackage{float}
\usepackage{amsmath}
\usepackage{listings}
\usepackage{xcolor}
\usepackage{subfig}
\usepackage{url}

\usepackage{tikz}
\usepackage{textcomp}
\usepackage{hyperref}
\usepackage{lipsum}

\newcommand\copyrighttext{%
  \footnotesize \textcopyright 2020 IEEE. Personal use of this material is permitted. 
    Permission from IEEE must be obtained for all other uses, in any current or future media, 
    including reprinting/republishing this material for advertising or promotional purposes, 
    creating new collective works, for resale or redistribution to servers or lists, 
    or reuse of any copyrighted component of this work in other works.
  DOI: \href{https://doi.org/10.1109/ICCCNT49239.2020.9225459}{10.1109/ICCCNT49239.2020.9225459}}
\newcommand\copyrightnotice{%
\begin{tikzpicture}[remember picture,overlay]
\node[anchor=south,yshift=10pt] at (current page.south) {\fbox{\parbox{\dimexpr\textwidth-\fboxsep-\fboxrule\relax}{\copyrighttext}}};
\end{tikzpicture}%
}

\title{Deep Learning based approach to detect Customer Age, Gender and Expression in Surveillance Video}
\author{
\IEEEauthorblockN{Dr. Earnest Paul Ijjina\IEEEauthorrefmark{1}}
\IEEEauthorblockA{Assistant Professor, Department of Computer Science and Engineering\\
	National Institute of Technology Warangal, India-506004\\
	Email : \IEEEauthorrefmark{1}iep@nitw.ac.in}

\IEEEauthorblockN{Goutham Kanahasabai\IEEEauthorrefmark{2}, Aniruddha Srinivas Joshi \IEEEauthorrefmark{3}}
\IEEEauthorblockA{B.Tech., Department of Computer Science and Engineering\\
	National Institute of Technology Warangal, India-506004\\
	Email : \IEEEauthorrefmark{2} gauthamkanags@gmail.com, \IEEEauthorrefmark{3} aniruddha980@gmail.com}
}

\begin{document}
\maketitle
\copyrightnotice

\begin{abstract}
In the current information era, customer analytics play a key role in the success of any business. Since customer demographics primarily dictate their preferences, identification and utilization of age \& gender information of customers in sales forecasting, may maximize retail sales. In this work, we propose a computer vision based approach to age and gender prediction in surveillance video. The proposed approach leverage the effectiveness of Wide Residual Networks and Xception deep learning models to predict age and gender demographics of the consumers. The proposed approach is designed to work with raw video captured in a typical CCTV video surveillance system. The effectiveness of the proposed approach is evaluated on real-life garment store surveillance video, which is captured by low resolution camera, under non-uniform illumination, with occlusions due to crowding, and environmental noise. The system can also detect customer facial expressions during purchase in addition to demographics, that can be utilized to devise effective marketing strategies for their customer base, to maximize sales.
\end{abstract}
\section{Introduction}
\label{intro}
In developing counties like India, where retail sales is major means of marketing, there is a huge potential in utilizing customer behaviour analysis to optimize sales. The customer demographics such as age and gender, in addition to the sentiment a customer experiences towards particular products plays a significant role in the sales and operations of retailers and small scale vendors. The analysis of the customer base is crucial for retailers to stay on top in the business marketing and maintaining a healthy group of consumers to make profits. As a retailer, anticipating and understanding the wants and interests of the existing customers can prove to be an effective route to tackle the battle of knowing what to stock up for the future. Such analysis can help retailers in ensuring that customers find what they are looking for, and can thus ensure a growing consumer base towards the business. Apprehending the aforementioned traits not only helps small scale vendors, but also advocates retail mall's such as Walmart and Target in better engaging customers \cite{ref1, ref2}. The knowledge of the average shopper in addition to the expressions one portrays while shopping, plays a paramount role.

In the consumer base, we find that each individual customer exhibit inclination towards a said commodity. So categorizing customers by age and gender is essential for better marketing. The studies \cite{ref3} suggest that around 20\% of new businesses survive their first year of operation and half of the small businesses close down in the first 5 years. Although several factors such as on-line competition, low profits, unavailable resources etc., may lead to the decline of a business. Analysing the consumer base over a period and understanding their sentiments over the product line-up can aid retailers in securing better profit margins ensuring survival and growth in the long run. Targeting consumers by segmenting the market ensures a healthy increase in consumers over time, which is a crucial factor in the functioning of current businesses.\par

Facial expression recognition has gained significant attention in the field of marketing, aiding enterprises to comprehend opinions expressed by customers while purchasing products. Traditionally, majority of enterprises used conventional methods of marketing such as advertisements, customer satisfaction surveys etc. These methods often prove to be rather time consuming and are more often than not an expensive venture. With the advent of technology, customer analytics can be used to deliver fitting products to the consumers. Sentiment is the emotion behind engaging consumers, and capturing customer sentiment helps us understand the following metrics that can aid in devising upcoming marketing strategies. Some of the common customer sentiments are:
\begin{itemize}
    \item \textbf{Overall satisfaction:} this helps us understand whether the experience at the location proved fruitful for the customer
    \item \textbf{Loyalty:} whether existing customers would recommend the said retailer to others 
    \item \textbf{Future Engagement:} whether the customers would engage with the retailer in the future
\end{itemize}
Analysing the aforementioned metrics gives valuable insight to potent marketing strategies that are more engaging and beneficial to the customer.

In this work, we propose a deep learning based framework to identify the age, gender and expressions of the customers from surveillance videos. Video surveillance through Closed Circuit TeleVision (CCTV) cameras is the most commonly used set-up to monitor and track humans in retail stores \cite{ref4}.

In this paper, we propose an approach to detect age, gender and expressions of the customers effectively in CCTV footage obtained from a typical video surveillance system. The surveillance videos from CCTV cameras in stores is given as input to the proposed framework to detect faces and estimate age, gender and expression of the person. The proposed framework is designed to operate under real world environmental conditions like poor illumination, over exposure of background, challenging angles of view etc., which makes the recognition/estimation task more challenging. 

\section{Related Works and Literature}
\label{literature}
In this work, we primarily focus on two pivotal demographics of the consumer base: age and gender, which are the crucial factors in helping retailers and marketers come up with effective business strategies. In addition, we also focus on extracting the facial expressions that the customers exhibit during their engagement in the stores, for reasons mentioned in section \ref{intro}. A typical computer vision approach for age detection either assigns a numerical age of the estimated age (or) a categorical age-group like youth, old etc., to each subject. The typical gender recognition system outputs a binary label for each subject indicating the gender as male (or) female.

The earliest work on Age classification involved a method proposed by Kwon \textit{et al.} that extracted the primary facial features such as eyes, chin, mouth, nose etc. and these features were therein used to compute ratios that set apart seniors from toddlers \cite{ref5}. The computed ratios aid a wrinkle index which is used to categorise the people into three different age groups. However, this approach has the constraint that the person in consideration needs to be looking directly at the camera and it does not provide us with age as a physical number. Moreover, this approach requires an accurate localisation of facial features, which is a challenging feat on its own and thus proves inadequate for real world environments such as shops. A similar approach to model the progression of age attributes was proposed by Ramanathan \textit{et al.} \cite{ref6} but it too suffers from the aforementioned limitations.

An aproach based on local features representing images of faces is proposed by Yan \textit{et al.}. This methodology involved the utilization of Gaussian Mixture Models (GMM) \cite{ref7} to symbolise the distribution of facial patches \cite{ref8}. In a similar work \cite{ref9}, GMMs were used for representing the distribution of facial measurements where more robust descriptors were used instead of the traditional pixel patches. Super vectors were used for representing facial patch distributions in \cite{ref10}.

The methods described above prove to be effective on datasets containing frontal images of faces and are not suitable for surveillance videos that always tend to have images of faces captured at different angles.\par
Coming to gender classification, early works include a method proposed by Moghaddem \textit{et al.} \cite{ref11} that involved the utilisation of Support Vector Machine(SVM) classifiers, wherein they were applied to input intensity values of images for classifying gender. Another work that involved image intensities was by Baluja \textit{et al.} \cite{ref12}, however they used Adaboost.

For the purpose of gender classification, most of the works tend to utilise a well versed dataset known as the FERET benchmark \cite{ref13}. The FERET dataset is a composition of images that were obtained from well illuminated, controlled environments and are thus not by any means close to images obtained from the frames of CCTV videos, which involves daunting factors such as noise, occlusions in between frames etc. As a result, we tend to focus more on CCTV videos placed in stores which tend to capture the environment in its natural element, posing credible challenges to work with.

Expression recognition from facial features has gain attention in the marketing space as it is a natural indicator of a customer’s emotions. In the work of Ekman \textit{et al.}, Facial Action Coding System(FACS) \cite{ref14}, action parameters were designated to each of the facial expressions which were therein used to classify human emotions.

The layout of the rest of the paper is as follows: Section \ref{approach} delineates the proposed framework of the paper. Section \ref{results} contains the analysis of our experimental results. Section \ref{conclusion} gives the conclusions and future work. Section \ref{ack} contains the acknowledgements.

\section{Proposed Approach}
\label{approach}
\begin{figure} 
	\centering
	\includegraphics[width=80mm, height=150mm]{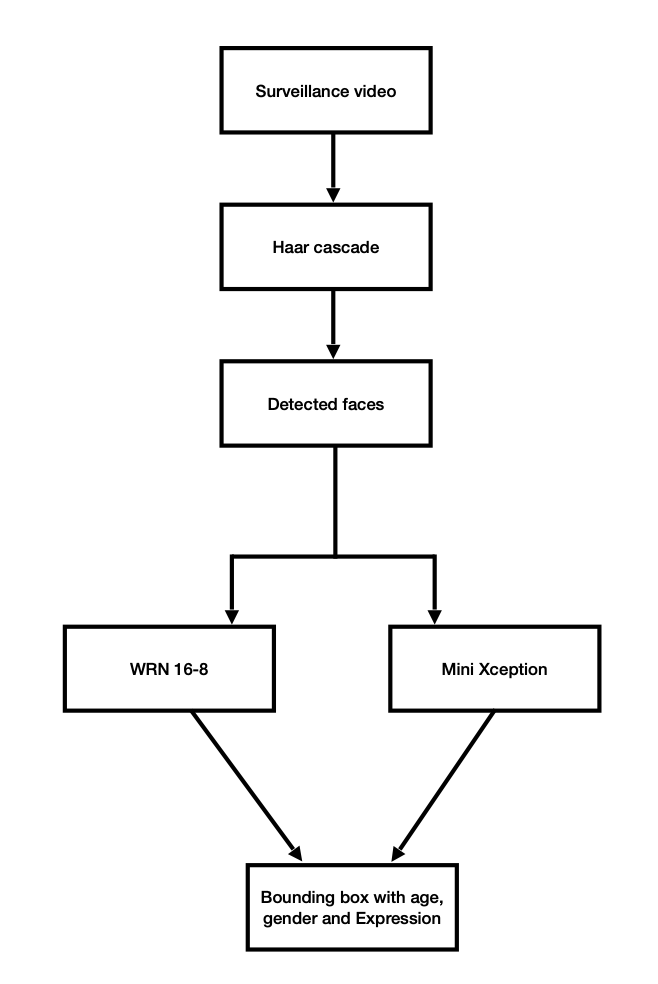}
	\caption{Workflow of the proposed Framework}
	\label{fig:workflow}
\end{figure}

In this section, we elaborate our proposed framework, which is illustrated in Figure \ref{fig:workflow}. We describe how the framework is realized in order to detect the customer related information. Our predominant goal is to identify the age, gender and expressions of customers in surveillance video, so that it can be further utilized in consumer analytics to anticipate sales. \par

The proposed methodology involves the following tasks:
\begin{enumerate}
    \item [T1:]  Face Detection
    \item [T2:]  Age and Gender Estimation
    \item [T3:]  Expression recognition
\end{enumerate}
The details of each of these tasks will be exemplified in the remainder of this section.

\subsection{Face Detection}
\label{facedetect}
In this section, we describe the method used to detect the region of the face in the video frames. For this purpose, we use the Haar Cascade object detection model \cite{ref15}, which is a machine learning based approach wherein a cascade function is trained on a positive and negative images, i.e., images with faces and images without faces respectively, thereby detecting faces like objects in given images.

The details of this method is as follows. The algorithm consists of four major steps, namely: Haar Feature Selection, Creating Integral Images, Adaboost Training and finally Cascading Classifiers. The first step is to collect all the Haar features, which operate on rectangular regions in the detection window, calculate the intensity of the pixels in each of these rectangular regions, enumerate the difference between these sums. Integral images are used to speed up this process. Among all the features extracted, the best and the most relevant features are selected using Adaboost, which performs the aforementioned task in addition to training the classifiers that use them. Adaboost builds a strong classifier as a linear weighted combination of weak classifiers. Each Haar feature acts like a weak classifier, and a significant amount of Haar features characterizing the object are cascaded to form a strong classifier.

The cascade classifier is therefore an agglomeration of stages, wherein each stage is a unity of weak learners. Stages are trained to high degrees of accuracy by taking into picture, a weighed average of the decisions undertaken by the weak learners. Each stage of the classifier labels the current region as either positive or negative, indicating that an object was found or not found respectively. The output of this stage is the coordinates of the bounding boxes enclosing the face regions in the input. This information is utilized in remaining tasks i.e, T2 and T3.

\subsection{Age and Gender Estimation}
\label{ageandgender}
\begin{figure}
	\centering
	\includegraphics[width=70mm, height=120mm]{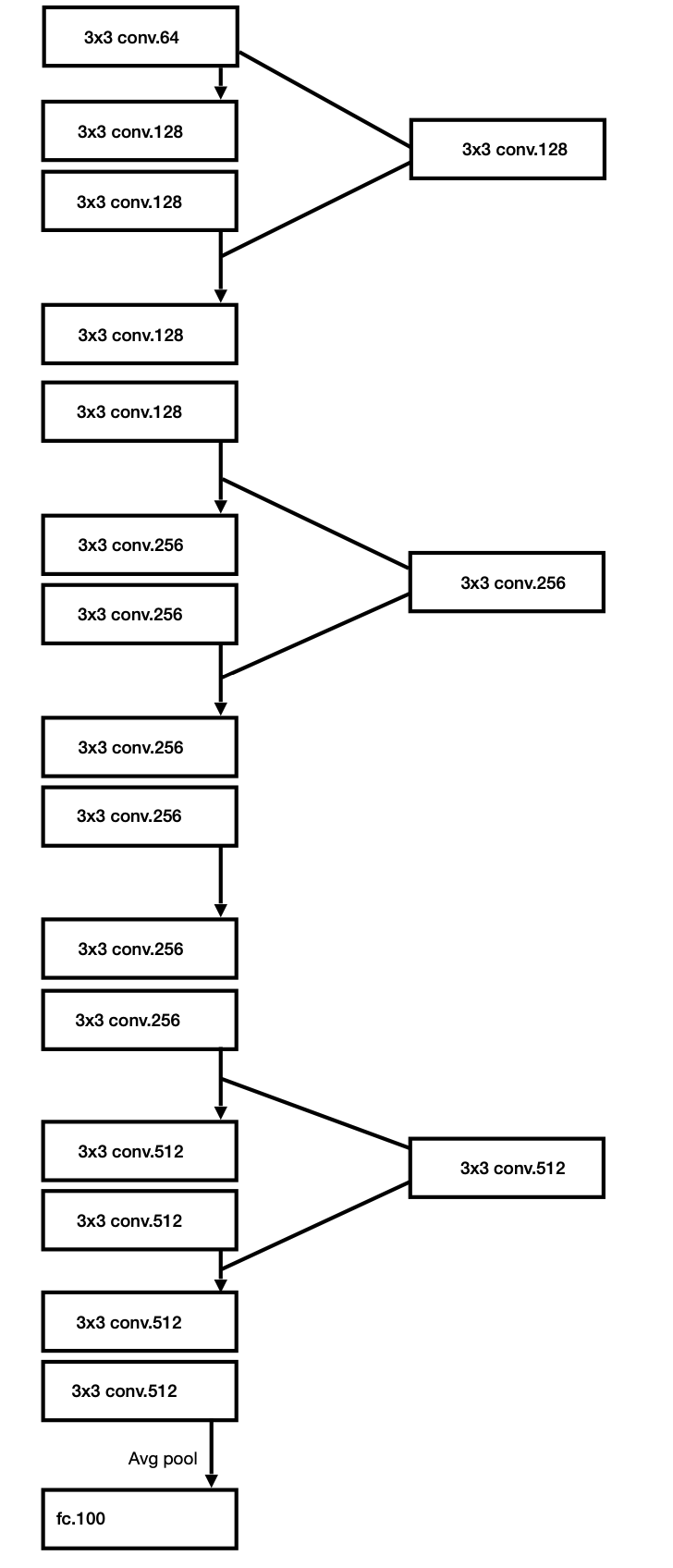}
	\caption{WRN-16-8 model architecture}
	\label{fig:WRN}
\end{figure}

In this section, we describe the methodology used to extract the age and gender corresponding to the face region obtained from task T1. In this task, we use the Wide ResNet 16-8 (WRN-16-8), a Wide Residual Network \cite{ref21} architecture to estimate age and gender. The architecture considered in this work gave preference to width over depth as effective training of deeper models is a complex task. We chose a Wide Residual Network architecture with more convolution layer filters to improve its effectiveness with less number of layers. The other key advantage of increasing width instead of depth is more computational efficiency. This Wide ResNet model when compared to the traditional ResNet \cite{ref21}, has less number of layers and performing twice as fast as the former. The model considered in this work, the Wide ResNet $16-8$ , i.e WRN-$16-8$ with 16 convolutional layers and a widening factor of 8, which is the number of feature maps per layer is shown in Figure \ref{fig:WRN}. The input to this network is a 64$\times$64 RGB image. This 16 layer neural network has the same accuracy as a deep network which has 1000-thin layers, and is also several times faster to train. This suggests that residual blocks are crucial factor to the effectiveness of a deep residual network. The output of this phase is the estimated age, which is numeric value and the gender i.e., male or female, of the subject.

\subsection{Expression recognition}
\label{emotionrecog}
In this section, we explain the methodology used for facial expression recognition. We utilize the output of task T1, which is the region of the face in the frame, enclosed by the bounding box. The model chosen for recognizing the facial expression is the mini Xception model \cite{ref19} which is inspired by the well versed Xception architecture \cite{ref16}. The mini Xception model doesn't use fully connected (FC) layers in the network architecture, but utilizes the residual modules \cite{ref17} and depth-wise separable convolutions \cite{ref18}. The use of depth-wise separable convolutions reduce the computation in comparison to regular convolutions. Residual models serve the role of altering the desired mapping between two successive layers in the network, causing the difference between the original and desired featured map to be learnt. 

\begin{figure}
	\centering
	\includegraphics[width=70mm, height=120mm]{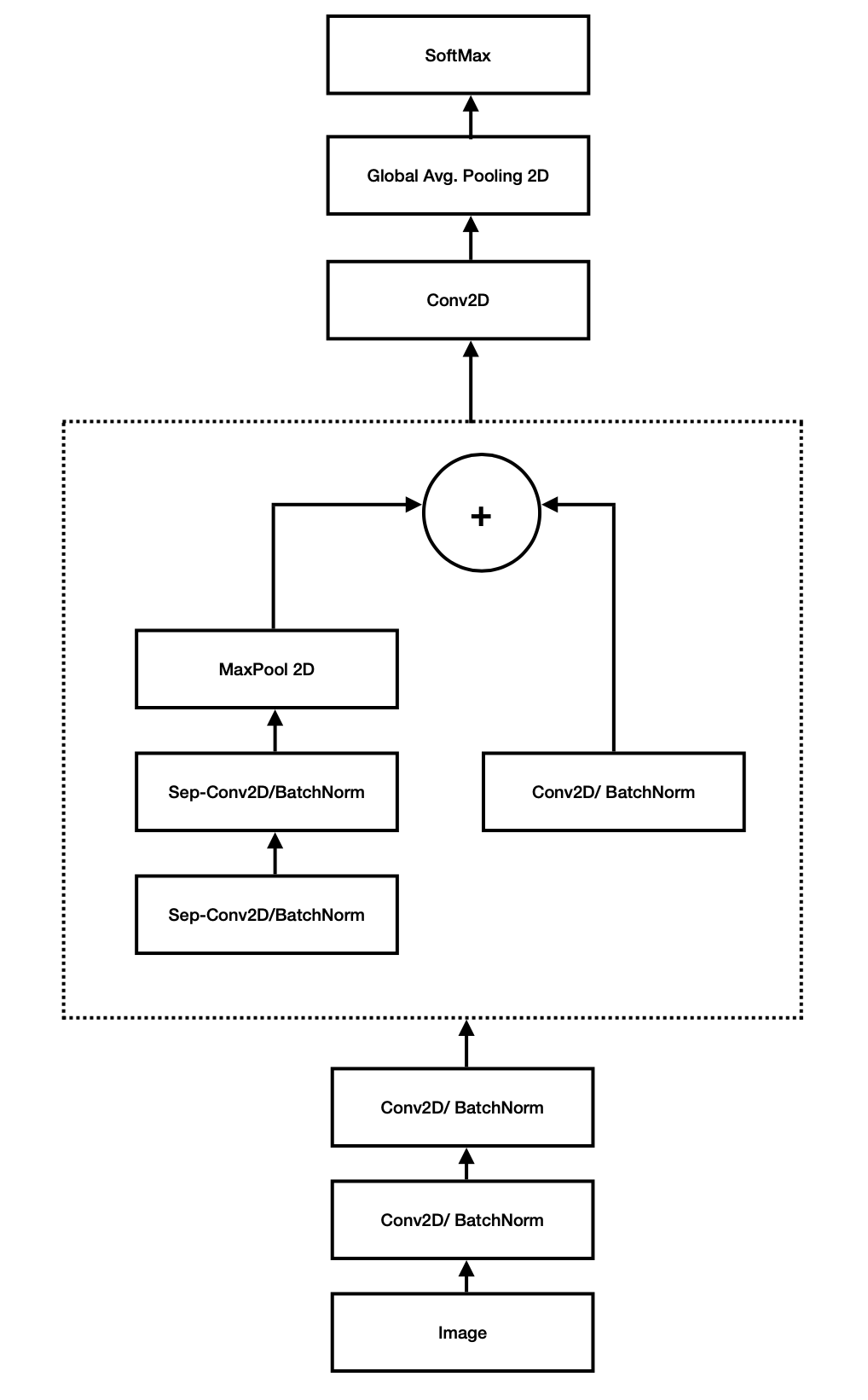}
	\caption{mini Xception model architecture}
	\label{fig:mini_xception}
\end{figure}

Figure \ref{fig:mini_xception} illustrates the mini Xception architecture considered in this work. It is a Fully Convolutional Network (FCN) with 4 residual depth wise separable convolutions, with each convolution operation being followed by a batch normalisation and ReLU activation. Global average pooling and a softmax activation function is applied in the last layer to predict the expression.

The output of this phase, is the expression portrayed by the person, which belongs to one of {\textit{happy}, \textit{sad}, \textit{anger}, \textit{neutral}, \textit{surprise}, \textit{fear}, \textit{disgust}}. This behavioural information can be used to understand the customer preferences of merchandise.

\begin{figure}
	\centering
	\includegraphics[width=90mm, height=120mm]{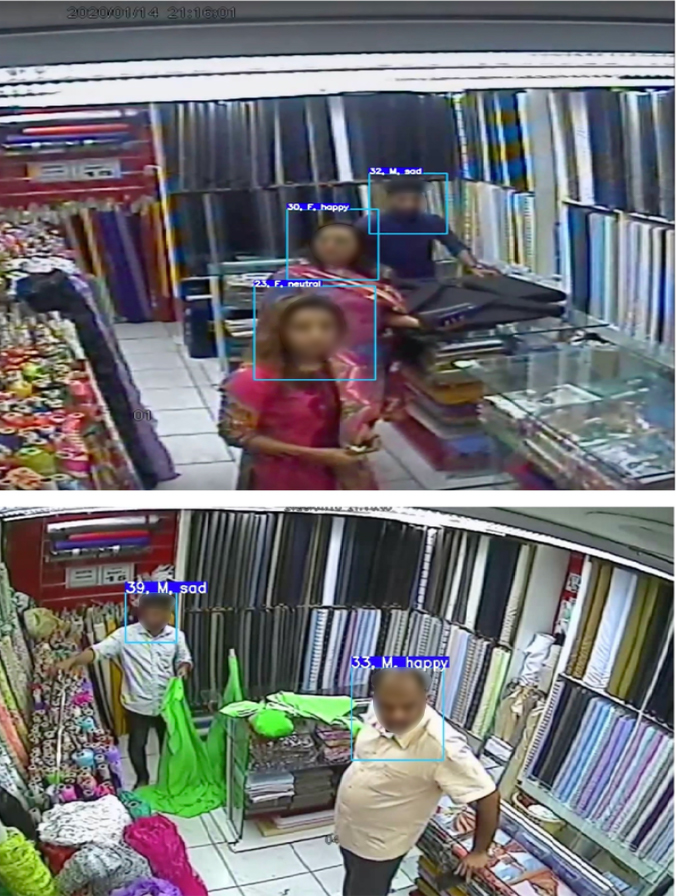}
	\caption{Visualization of results for video}
	\label{fig:results}
\end{figure}

\section{Experimental Evaluation}
\label{results}
In this section, we discuss the results of our proposed approach on the surveillance video dataset. All the experiments in this work were conducted on Google CoLab \cite{refx3} in an environment with Intel(R) Xeon(R) CPU @ 2.00GHz with NVIDIA Tesla T4 GPU, 16 GB GDDR6 VRAM and 13GB RAM. All programs were written in Python - 3.6 and utilized Keras - 2.3.1 and OpenCV - 4.2.0.

\subsection{Dataset Used}
\begin{figure*}
	\centering
	\includegraphics[width=180mm, height=160mm]{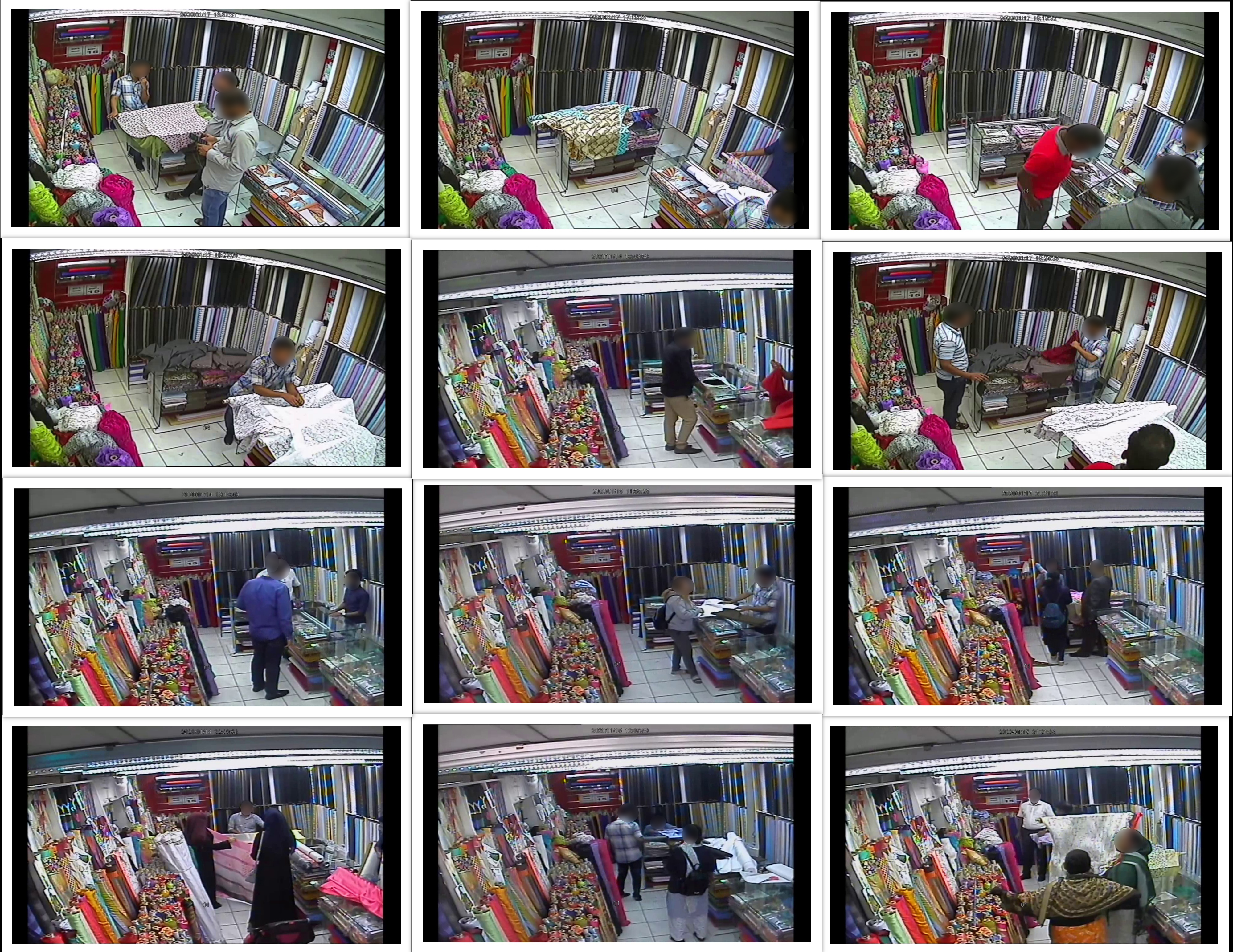}
	\caption{Sample video in the dataset}
	\label{fig:dataset}
\end{figure*}

The dataset used in this study is a collection of surveillance videos of a garment store, capturing sales information. The videos contain the interaction between the salesmen and customers while selecting a garment for purchase. The videos are obtained from 2 different CCTV cameras in the store's infrastructure, thus enabling us to work on real-life surveillance videos taken from multiple angles of view and practical illumination conditions. There are a total of 15 video samples in the dataset with an average running time of 5 minutes. The resolution of the videos is 944 $\times$ 576 pixels. The raw videos without any preprocessing like enhancement (or) de-noising is used in this work to evaluate the suitability of the model for practical use. The typical sales videos in the dataset are shown in Figure \ref{fig:dataset}.

\subsection{Results and Statistics}

This section presents the results obtained by evaluating the proposed approach on the video dataset. Figure \ref{fig:results} illustrates the output of the proposed approach on few instances of the dataset. The bounding box around the detected faces as described in Section \ref{facedetect} for task T1, which is then utilized to obtain age, gender and facial expression as described in Section \ref{ageandgender} for task T2 and Section \ref{emotionrecog} for task T3 respectively.

The numerical results of the proposed approach are given in Table \ref{tab:my-table}. On the surveillance videos dataset, the model achieved a Gender classification accuracy of 82.9\%.  The age estimation is evaluated to assign the subject to the right age group, in \{0-9, 10-19,....,70-79\} ranges. For this metric, the model obtained an accuracy of 70.8\%. These results were obtained on real-world in-door video with significant background noise.

\begin{table}[h]
\centering
\renewcommand{\arraystretch}{1.5}
\begin{tabular}{|l|l|l|}
\hline
Proposed  & Gender & Age\\ \cline{2-3} 
approach &  82.926 & 70.804 \\ \hline
\end{tabular}
\caption{Performance of the proposed model in percentage}
\label{tab:my-table}
\end{table}

The model achieved better accuracy when the subjects is facing the camera. However, its accuracy reduces when the customer is not facing the camera directly. The framework still managed to achieve good estimate of age in both scenarios.

The model had difficulties in detecting the side facial profile of customers not facing the camera, thereby affecting its accuracy. However, the framework still managed to achieve good estimate of age in such scenarios.

The predicted expressions were manually verified, as the labels for of the expressions of the subjects in the surveillance video is unavailable i.e., unlabelled video. The manual verification suggests that the expressions were reasonably accurate. Similar works in the field of face and gender detection do not cater specifically to identifying subjects of Indian origin, they were largely trained on established datasets comprising of facial images of people from the west.  This model was optimized for detecting the age and gender traits of customers belonging to Indian origin. 

\section{Conclusion and Future Works}
\label{conclusion}
In this work, an approach to detect the age, the gender and the expressions of the consumers in surveillance video is proposed. The proposed approach is able to recognize faces of consumers in below average video resolution and is also able to identify their age and gender to a fair degree of accuracy, achieving a 82.9\% accuracy rate for gender and 70.8\% accuracy for age-range. The accuracy of the predicted expressions is verified manually, and is found to be reasonably accurate. The future work aims to extend this work to more challenging data.
\section{Acknowledgement}
\label{ack}
This work was done by Mr. Goutham Kanahasabai and Mr. Aniruddha Srinivas Joshi (final year B.Tech students) under the guidance of Dr. Earnest Paul Ijjina (Assistant Professor), in Department of Computer Science and Engineering, National Institute of Technology Warangal, as a part of their final year project. The authors express their gratitude to the CSE department and the Institute, NIT Warangal for their efforts in developing the research environment, where this study was conducted. We also thank Google Colab for providing access to computational resources used to run these algorithms.
\bibliographystyle{IEEEtran}
\bibliography{main}

\begin{thebibliography}{10}
\providecommand{\url}[1]{#1}
\csname url@samestyle\endcsname
\providecommand{\newblock}{\relax}
\providecommand{\bibinfo}[2]{#2}
\providecommand{\BIBentrySTDinterwordspacing}{\spaceskip=0pt\relax}
\providecommand{\BIBentryALTinterwordstretchfactor}{4}
\providecommand{\BIBentryALTinterwordspacing}{\spaceskip=\fontdimen2\font plus
\BIBentryALTinterwordstretchfactor\fontdimen3\font minus \fontdimen4\font\relax}
\providecommand{\BIBforeignlanguage}[2]{{%
\expandafter\ifx\csname l@#1\endcsname\relax
\typeout{** WARNING: IEEEtran.bst: No hyphenation pattern has been}%
\typeout{** loaded for the language `#1'. Using the pattern for}%
\typeout{** the default language instead.}%
\else
\language=\csname l@#1\endcsname
\fi
#2}}
\providecommand{\BIBdecl}{\relax}
\BIBdecl

\bibitem{ref1}
``Walmart customer demographics,'' \url{https://snapshot.numerator.com/retailer/walmart}.

\bibitem{ref2}
``The average target shopper,'' https://extension.psu.edu/understanding-your-customers-how-demographics-and-psychographics-can-help.

\bibitem{ref3}
https://www.shopkeep.com/blog/why-small-businesses-fail.

\bibitem{ref4}
T.~Kanade, R.~Collins, A.~Lipton, H.~Fujiyoshi, and D.~Duggins, ``A system for video surveillance and monitoring cmu vsam final report,'' p. 135, 11 1999.

\bibitem{ref5}
Y.~H. Kwon and N.~da~Vitoria~Lobo, ``Age classification from facial images.''

\bibitem{ref6}
N.~{Ramanathan} and R.~{Chellappa}, ``Modeling age progression in young faces,'' in \emph{Proc. of IEEE Computer Society Conference on Computer Vision and Pattern Recognition (CVPR'06)}, vol.~1, June 2006, pp. 387--394.

\bibitem{ref7}
G.~{Guo}, Y.~{Fu}, C.~R. {Dyer}, and T.~S. {Huang}, ``Image-based human age estimation by manifold learning and locally adjusted robust regression,'' \emph{IEEE Transactions on Image Processing}, vol.~17, no.~7, pp. 1178--1188, July 2008.

\bibitem{ref8}
K.~Fukunaga, \emph{Introduction to Statistical Pattern Recognition (2nd Ed.)}.\hskip 1em plus 0.5em minus 0.4em\relax USA: Academic Press Professional, Inc., 1990.

\bibitem{ref9}
{Shuicheng Yan}, {Xi Zhou}, {Ming Liu}, M.~{Hasegawa-Johnson}, and T.~S. {Huang}, ``Regression from patch-kernel,'' in \emph{Proc. of IEEE Conference on Computer Vision and Pattern Recognition}, June 2008, pp. 1--8.

\bibitem{ref10}
S.~Yan, M.~Liu, and T.~Huang, ``\BIBforeignlanguage{English (US)}{Extracting age information from local spatially flexible patches},'' in \emph{\BIBforeignlanguage{English (US)}{Proc. of IEEE International Conference on Acoustics, Speech and Signal Processing (ICASSP)}}, ser. Proc. of IEEE International Conference on Acoustics, Speech and Signal Processing (ICASSP), 9 2008, pp. 737--740.

\bibitem{ref11}
B.~{Moghaddam} and {Ming-Hsuan Yang}, ``Learning gender with support faces,'' \emph{IEEE Transactions on Pattern Analysis and Machine Intelligence}, vol.~24, no.~5, pp. 707--711, May 2002.

\bibitem{ref12}
S.~Baluja and H.~Rowley, ``Boosting sex identification performance,'' \emph{International Journal of Computer Vision}, vol.~71, pp. 111--119, 06 2007.

\bibitem{ref13}
P.~J. {Phillips}, {Hyeonjoon Moon}, S.~A. {Rizvi}, and P.~J. {Rauss}, ``The feret evaluation methodology for face-recognition algorithms,'' \emph{Proc. of IEEE Transactions on Pattern Analysis and Machine Intelligence (PAMI)}, vol.~22, no.~10, pp. 1090--1104, Oct 2000.

\bibitem{ref14}
P.~Ekman and W.~V. Friesen, ``Facial action coding system: Manual,'' 1978.

\bibitem{ref15}
P.~{Viola} and M.~{Jones}, ``Rapid object detection using a boosted cascade of simple features,'' in \emph{Proc. of IEEE Computer Society Conference on Computer Vision and Pattern Recognition (CVPR)}, vol.~1, Dec 2001, pp. I--I.

\bibitem{ref21}
S.~Zagoruyko and N.~Komodakis, ``Wide residual networks,'' 2016.

\bibitem{ref19}
O.~Arriaga, M.~Valdenegro, and P.~Plöger, ``Real-time convolutional neural networks for emotion and gender classification,'' 10 2017.

\bibitem{ref16}
F.~{Chollet}, ``Xception: Deep learning with depthwise separable convolutions,'' in \emph{Proc. of IEEE Conference on Computer Vision and Pattern Recognition (CVPR)}, 2017, pp. 1800--1807.

\bibitem{ref17}
K.~{He}, X.~{Zhang}, S.~{Ren}, and J.~{Sun}, ``Deep residual learning for image recognition,'' in \emph{Proc. of IEEE Conference on Computer Vision and Pattern Recognition (CVPR)}, 2016, pp. 770--778.

\bibitem{ref18}
A.~G. Howard, M.~Zhu, B.~Chen, D.~Kalenichenko, W.~Wang, T.~Weyand, M.~Andreetto, and H.~Adam, ``Mobilenets: Efficient convolutional neural networks for mobile vision applications,'' 2017.

\bibitem{refx3}
``Google colab,'' \url{https://colab.research.google.com/}.

\end{thebibliography}
\end{document}